\documentclass[letterpaper]{article} 
\usepackage{aaai2026}  
\usepackage{times}  
\usepackage{helvet}  
\usepackage{courier}  
\usepackage[hyphens]{url}  
\usepackage{graphicx} 
\usepackage{natbib}  
\usepackage{caption} 
\usepackage{algorithm}
\usepackage{algorithmic}
\usepackage{booktabs}

\usepackage{newfloat}
\usepackage{listings}
\DeclareCaptionStyle{ruled}{labelfont=normalfont,labelsep=colon,strut=off} 
\floatstyle{ruled}
\newfloat{listing}{tb}{lst}{}
\floatname{listing}{Listing}
\nocopyright

\title{Metaphors We Compute By: A Computational Audit of \\ Cultural \emph{Translation} vs. \emph{Thinking} in LLMs}
\author {
    Yuan Chang\thanks{Not reflecting the authors' positions at Meta/Amazon.}\equalcontrib\textsuperscript{\rm 1},
    Jiaming Qu\footnotemark[1]\footnotemark[2]\textsuperscript{\rm 2},
    Zhu Li\footnotemark[1]\textsuperscript{\rm 1}
}
\affiliations{
    \textsuperscript{\rm 1}Meta, Burlingame, CA\\
    \textsuperscript{\rm 2}Amazon, Seattle, WA\\
    yuanchang96@meta.com, qjiaming@amazon.com, zhuli@meta.com
}

\begin{document}

\maketitle
\begin{center}
\textit{Author-prepared preprint. Accepted to AAAI 2026 Spring Symposium.}
\end{center}

\begin{abstract}
Large language models (LLMs) are often described as multilingual because they can understand and respond in many languages. However, speaking a language is not the same as reasoning within a culture. This distinction motivates a critical question: do LLMs truly conduct culture-aware reasoning? This paper presents a preliminary computational audit of cultural inclusivity in a creative writing task. We empirically examine whether LLMs act as culturally diverse creative partners or merely as cultural translators that leverage a dominant conceptual framework with localized expressions. Using a metaphor generation task spanning five cultural settings and several abstract concepts as a case study, we find that the model exhibits stereotyped metaphor usage for certain settings, as well as Western defaultism. These findings suggest that merely prompting an LLM with a cultural identity does not guarantee culturally grounded reasoning.
\end{abstract}

\section{Introduction}

Large language models (LLMs) are often described as global technologies because they support many languages and are widely deployed across regions and cultures. This framing implicitly assumes that linguistic coverage implies cultural reasoning. However, speaking a language is not equivalent to reasoning within a cultural worldview~\cite{hill1988language}. Culture shapes how abstract concepts are structured, related, and expressed. As a result, translating culturally marked language does not guarantee that underlying cultural perspectives are preserved.

This motivates a fundamental question: when LLMs generate culturally conditioned content, do they reason within diverse cultural frameworks, or do they translate a dominant worldview into culturally flavored surface forms? In this work, we conduct a preliminary analysis using \textbf{metaphor generation}. Prior work in cognitive linguistics shows that metaphors play a central role in structuring reasoning about abstract concepts and provide a useful probe for cultural conceptualization~\cite{lakoff1980metaphors}.

While prior studies have curated benchmarks to evaluate LLMs’ cultural understanding through question-answering tasks~\cite{chiu-etal-2025-culturalbench,singh-etal-2025-global}, fewer have examined cultural bias in the context of creative writing. We argue that creative writing provides a meaningful setting for evaluating cultural awareness in LLMs. In such tasks, a model may vary its phrasing while still relying on a narrow and culturally dominant semantic structure~\cite{khan2025randomness}. Producing culturally appropriate creative content therefore requires more than surface-level adaptation; it requires sensitivity to culturally grounded ways of expressing meaning.

In this work, we adopt a geometric perspective on cultural reasoning by analyzing culturally conditioned metaphor generation in embedding space. We generate metaphors for several abstract concepts across multiple cultural settings and examine these outputs in terms of intra-cultural semantic diversity, conceptual clustering, and cultural defaultism. Our preliminary results show that (1) the model exhibits representational collapse for some culture-concept pairs, sometimes repeatedly generating near-identical metaphors, and (2) for several concepts, the culture-agnostic condition aligns more closely with the U.S. condition than with other cultures, suggesting a culturally non-neutral baseline. Together, these findings indicate that current LLMs may \emph{not} reliably function as culturally pluralistic creative partners, but often behave as cultural translators that re-skin a dominant conceptual framework with localized expressions.

In summary, we present a preliminary study for auditing cultural reasoning in LLMs. We introduce computational approaches for evaluating LLM's cultural reasoning and provide empirical evidence that cultural persona prompting alone is insufficient for ensuring culturally inclusive reasoning. By examining how LLMs generate metaphors across cultures, this work contributes to the symposium’s discussion on how AI systems can support co-creativity across diverse cultural contexts.
\section{Methods}

\subsection{Task Overview} We operationalize cultural reasoning in LLMs through the task of metaphor generation. Metaphors encode culturally grounded mappings between abstract concepts and lived experience, and prior work suggests they structure how people reason about concepts such as time, success, freedom, and death~\cite{lakoff1980metaphors}. We therefore use metaphor generation as a diagnostic probe of whether an LLM represents distinct cultural worldviews or merely adapts surface-level phrasing.

For this preliminary study, we focused on five abstract concepts: \textit{Time}, \textit{Death}, \textit{Success}, \textit{Family}, and \textit{Freedom}. These concepts are culturally salient and known to exhibit cross-cultural variation in interpretation. We experimented with five cultural settings: \textit{United States (U.S.)}, \textit{Japan}, \textit{China}, \textit{India}, and \textit{Brazil}, as well as a culture-unspecified condition denoted as \textit{Default}. This yields 6 conditions in total (5 specific cultures + the default condition).

\subsection{Metaphor Generation}
For each \textit{(concept, culture)} pair, we prompted the model to generate a metaphor using the template:
\textit{``Generate a culturally grounded metaphor associated with [Culture] and complete this sentence: `[Concept] is like ...'. Avoid tourist clichés and shallow stereotypes.''}
Here, \textit{[Culture]} and \textit{[Concept]} are filled with one of the chosen cultures and concepts. All authors discussed and iteratively refined the prompt wording to encourage culturally grounded but non-stereotypical responses. Importantly, no cultural context is specified in the prompt under the \textit{Default} condition. To isolate conceptual variation from translation artifacts, we required all outputs to be in English regardless of culture. This ensures we compare metaphors on semantic content rather than language differences.

We used the \texttt{gemini-3-flash-preview} model via API with default generation settings. For each \textit{(concept, culture)} pair, we sampled 20 independent outputs, yielding 600 metaphors in total (5 concepts $\times$ 6 conditions $\times$ 20 runs).
To quantitatively analyze the generated metaphors, we embedded each sentence with a generated metaphor using \texttt{gemini-embedding-001}, which produces a 3072-dimensional sentence embedding. This maps each metaphor to a high-dimensional vector space. We then measured pairwise semantic similarity between metaphors using cosine distance (i.e., one minus cosine similarity). This metric ranges from 0 to 2, where 0 indicates that two sentences are semantically identical and 2 indicates that they are maximally dissimilar.

\subsection{Evaluations}
We conducted three analyses to assess whether the LLM-generated metaphors are culturally inclusive.

\textbf{Analysis I (Intra-Cultural Semantic Diversity):} For each (concept, culture) pair, we computed the average pairwise cosine distance among the 20 metaphor embeddings. This measures how diverse the model's metaphors are for a given concept within a single cultural condition. A higher average distance indicates a wider range of metaphors (greater creativity), whereas a low distance indicates that the model repeated very similar metaphors (potential representational collapse).

\textbf{Analysis II (Conceptual Space Geometry):} To assess how different abstract concepts relate to one another within each cultural condition, we examined the geometry of concept embeddings. We used t-SNE~\cite{vandermaaten2008tsne} to visualize all metaphor embeddings within a culture. This reveals how distinctly the model separates concepts (e.g., \textit{Time} vs. \textit{Family}) under a given cultural prompt, and whether cultural context causes certain concepts to be unduly conflated or separated.

\textbf{Analysis III (Cultural Defaultism Test):} We examined whether the \textit{Default} condition functions as a culturally neutral baseline. For each abstract concept, we computed the cosine distance between the centroid of the \textit{Default} cluster and the centroid of each culture-specific cluster. We then compared the distance between \textit{Default} and U.S. with the corresponding distances between \textit{Default} and other cultures. To assess statistical significance, we applied a one-sided Fisher randomization test~\cite{fisher1935design} with the null hypothesis that the \textit{Default} condition is equidistant from all cultural conditions. A significant result---where \textit{Default} is closer to U.S. than to another culture---indicates that the culture-agnostic prompt aligns disproportionately with a specific cultural worldview, which we refer to as \textit{cultural defaultism}.
\section{Results}

\subsection{Analysis I: Intra-Cultural Diversity}\label{results1}
\begin{figure}[htbp!]
\centering
\includegraphics[width=\columnwidth]{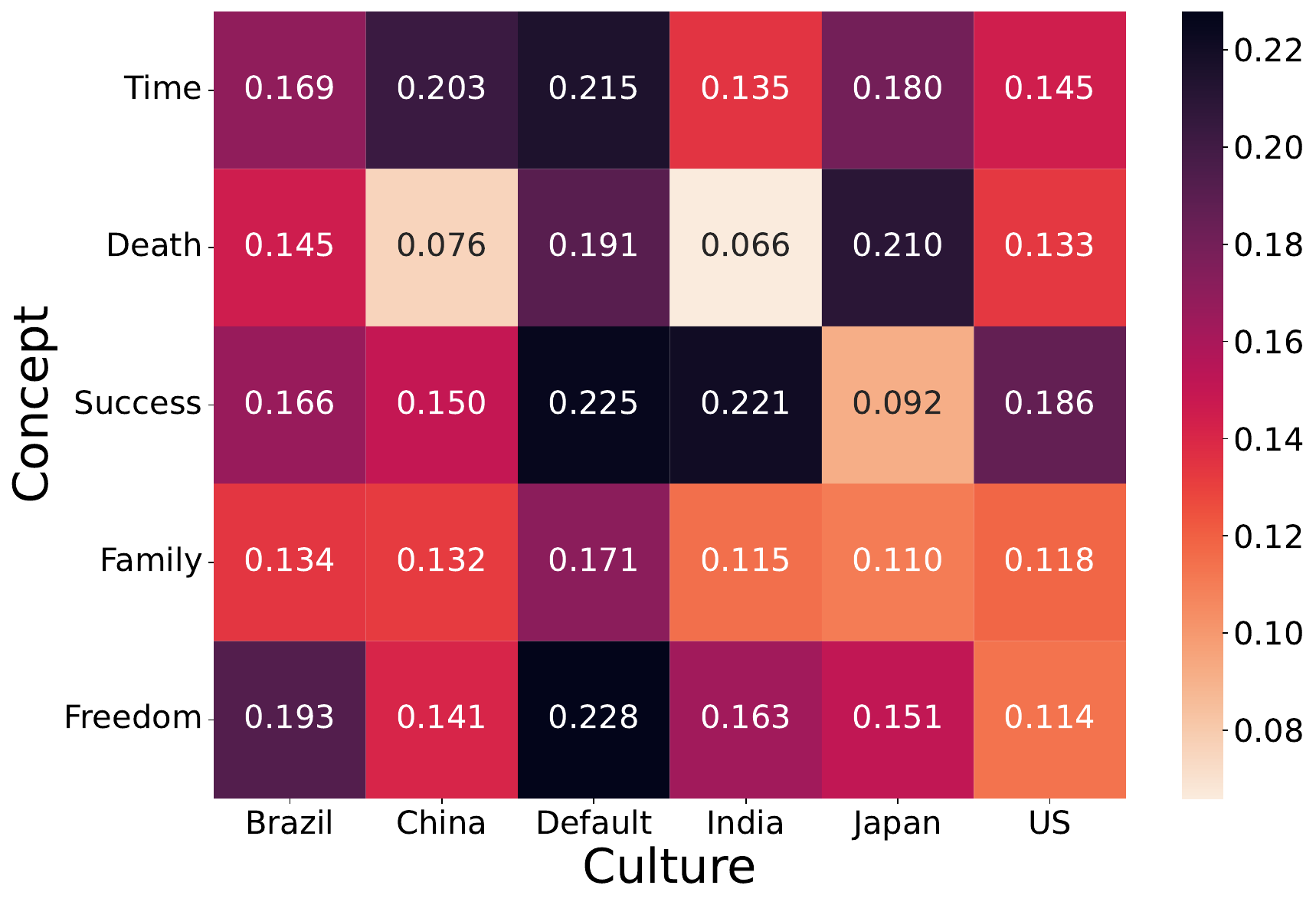} 
\vspace{-.4cm}
\caption{
\textbf{Intra-cultural semantic diversity of metaphors.}
Each cell shows the average pairwise cosine distance among 20 generated metaphors for a given (concept, culture) pair. Higher values indicate greater semantic diversity, while lower values indicate representational collapse.}
\label{fig:diversity}
\end{figure}

Figure~\ref{fig:diversity} shows a heatmap of the intra-cultural semantic diversity (i.e., average pairwise cosine distance within 20 runs) for every culture-concept pair. We observed two trends.

First, we observe substantial variation in diversity across both cultures and concepts. Some pairs exhibit very low diversity, indicating strong representational collapse. For example, metaphors for \textit{Death} in the India (0.066) and China (0.076) conditions form highly compact clusters. This means the model repeatedly relied on essentially the same metaphor rather than exploring a broad semantic space. In contrast, other pairs show relatively high diversity (e.g., \textit{Freedom} in \textit{Default} (0.228)), which indicates a wider range of metaphorical mappings for those cases.

Second, this collapse is asymmetric and concept-dependent. No single culture exhibits uniformly low or high diversity across all concepts. Instead, representational collapse occurs for specific concept–culture settings. For instance, the U.S. and \textit{Default} conditions have fairly high diversity for most concepts, yet even the \textit{Default} prompt shows lower diversity on certain concepts than some culturally prompted conditions.
Notably, the \textit{Default} condition often exhibits higher diversity than some culture-specific settings. This suggests that in some cases adding a cultural persona \textit{constrains} rather than enriches the model's representational space, resulting in a narrower set of metaphors.

\subsection{Analysis II: Conceptual Geometry}\label{results2}
\begin{figure}[htbp!]
\centering
\includegraphics[width=\columnwidth]{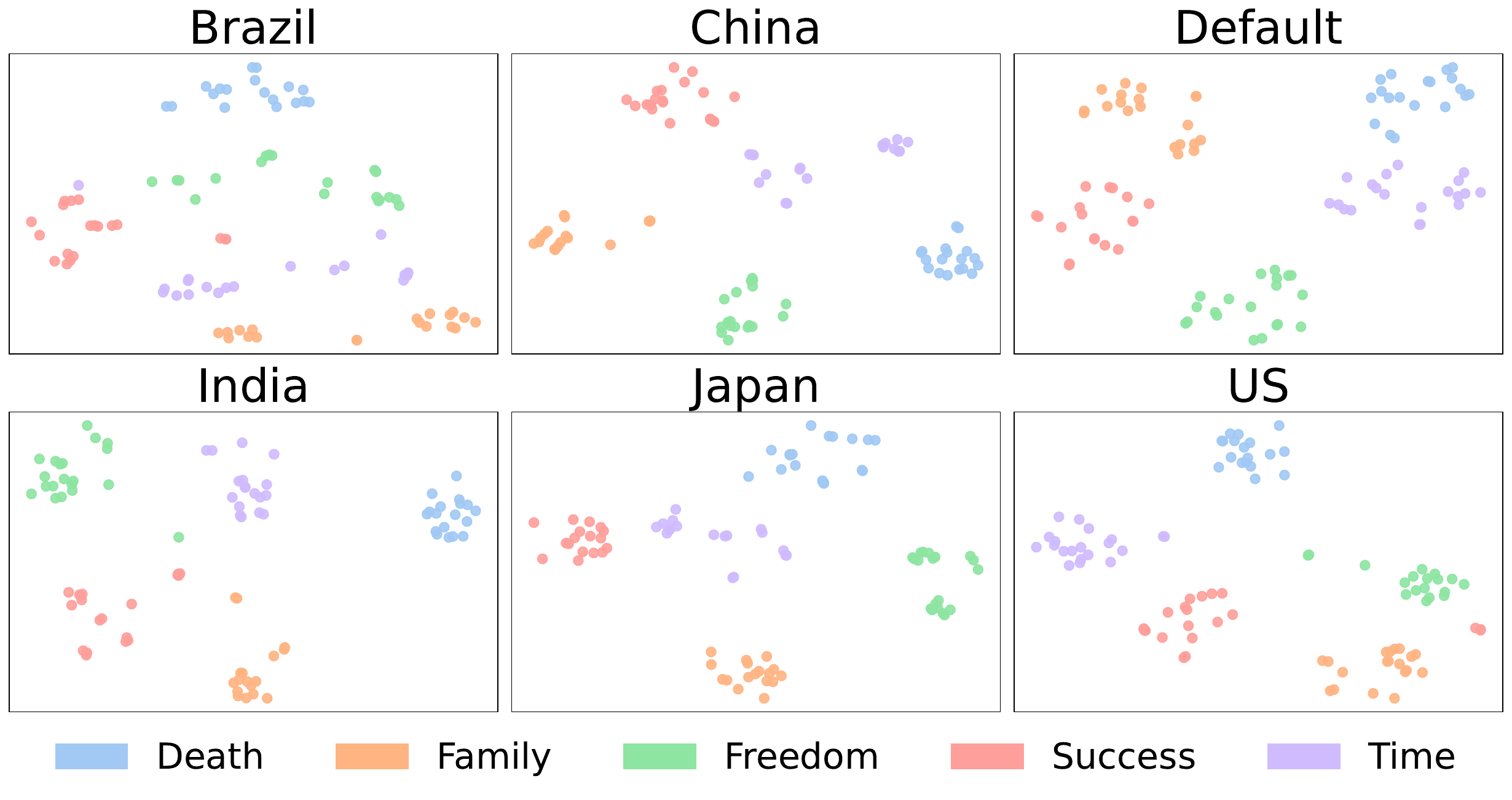} 
\vspace{-.4cm}
\caption{
\textbf{Conceptual geometry of metaphor embeddings across cultures.} Each panel shows a t-SNE projection of metaphor embeddings for one cultural condition. Colors indicate distinct abstract concepts.}
\label{fig:tsne}
\end{figure}

Next, we examine how cultural prompting reshapes the \emph{global organization} of abstract concepts in the model's embedding space. While the analysis above focuses on intra-cultural diversity within a single concept, this analysis provides a holistic view of how multiple concepts are positioned relative to one another under each cultural condition. Figure~\ref{fig:tsne} presents t-SNE visualizations of all metaphor embeddings for each culture.

Across cultural conditions, the overall configuration of concept clusters differs noticeably. In the \textit{Default} and \textit{Brazil} conditions, concept clusters are more broadly distributed across the embedding space, with larger gaps between clusters. In contrast, the \textit{China}, \textit{India}, \textit{Japan}, and \textit{U.S.} conditions exhibit more compact configurations, where multiple concepts occupy a relatively confined region of the projection. This suggests that cultural prompting can influence not only variation within concepts (Analysis I), but also the overall scale and spread of the model's conceptual organization.

At the same time, the visualizations reveal recurring patterns of conceptual proximity. Certain concept pairs, such as \textit{Family} and \textit{Success}, appear relatively close in several cultural conditions (e.g., \textit{Default}, \textit{U.S.}, and \textit{China}), while other concepts, such as \textit{Freedom}, are more consistently separated from the rest in multiple cultures. These tendencies suggest that some abstract relationships may remain stable across cultural prompts, whereas others are more sensitive to cultural context.

\subsection{Analysis III: Western Defaultism}\label{results3}
Finally, we tested whether the Default condition behaves as culturally neutral or if it implicitly leans Western. Table~\ref{tab:defaultism} reports the results of a permutation test comparing the distance between the Default and U.S. conditions against the distance between the Default and each non-U.S. culture.

\begin{table}[htbp!]
\centering
\small
\begin{tabular}{lcccc}
\toprule
\textbf{Concept} &
\textbf{Japan} &
\textbf{China} &
\textbf{India} &
\textbf{Brazil} \\
\midrule
Time     & -- & -- & $\uparrow^{**}$ & $\uparrow^{***}$ \\
Death    & -- & $\uparrow^{***}$ & $\uparrow^{***}$ & $\uparrow^{***}$ \\
Success  & $\downarrow^{*}$ & -- & -- & -- \\
Family   & -- & -- & -- & -- \\
Freedom  & -- & -- & $\uparrow^{*}$ & $\downarrow^{*}$ \\
\bottomrule
\end{tabular}
\vspace{-.1cm}
\caption{
\textbf{Directional results of the Western defaultism test.} Arrows indicate the direction of statistically significant differences in cosine distance between the Default condition and the US condition versus another culture.
$\uparrow$ indicates that Default is significantly closer to U.S. than to the corresponding culture (evidence of Western defaultism), while $\downarrow$ indicates that Default is significantly closer to the corresponding culture than to U.S. We used Fisher randomization test for significance testing: $^{*}p<0.05$, $^{**}p<0.01$, $^{***}p<0.001$.
}
\label{tab:defaultism}
\end{table}

We found statistically significant evidence of cultural defaultism for several concepts. For \textit{Time}, the \textit{Default} condition's embeddings are significantly closer to the U.S. embeddings than to those of India ($p < 0.01$) or Brazil ($p < 0.001$). For \textit{Death}, \textit{Default} is significantly closer to U.S. than to China, India, or Brazil (all $p < 0.001$). For \textit{Freedom}, the \textit{Default} condition is significantly closer to U.S. than to India ($p < 0.05$). These differences indicate that in these cases the ``culture-agnostic'' prompt yields metaphors that occupy a semantic position much nearer to the Western cultural conception than to other cultures' conceptions.

Notably, this effect is not universal across all concepts. For \textit{Family}, none of the cultural vs. U.S. distance comparisons were statistically significant. For \textit{Success} and \textit{Freedom}, we also observe cases where the Default condition is significantly closer to a non-U.S. culture than to the U.S. condition. These results indicate that the Default setting does not consistently privilege a single cultural worldview. Instead, its alignment varies by concept, sometimes reflecting non-U.S. cultural representations.

\subsection{Qualitative Analysis of Metaphors} To complement the above quantitative analyses, we also examined the generated metaphors themselves for recurring patterns or cultural tropes. We found that in many of the low-diversity cases, the model was indeed repeating a very similar metaphor that corresponds to well-known cultural idioms or archetypal metaphors. For example, in the \textit{India–Death} condition, 17 out of 20 generated metaphors invoked the same analogy: a clay pot shattering and its inner air merging back into the sky. This metaphor is explicitly rooted in the Advaita Vedanta concept of ``Ghata–Akasha''. Similarly, for \textit{China–Death}, the vast majority of outputs described ``a fallen leaf returning to its roots,'' reflecting an ancient Chinese idiom. These cases indicate the model knows a culturally salient metaphor and leans on it exclusively.

On the other hand, some culture-concept settings with higher semantic diversity exhibit a wider range of metaphors without a dominant recurring image. For instance, \textit{Success–India} outputs included metaphors ranging from stepwells, to hand-loomed textiles, to musical raga training, with no single theme dominating the runs. This indicates that when not constrained by a strong known cultural trope (or when the model has multiple examples to draw on), it can generate more varied and creative analogies. Similarly, the Default condition's outputs were often poetic but not explicitly tied to any one culture's symbols.
\section{Discussion and Conclusion}
\subsection{Summary of Findings}
In this work, we conducted a preliminary empirical audit of LLMs' ability to generate metaphors under different cultural prompts. We found three key trends. 
First, the model often exhibits representational collapse for certain culture-concept combinations (Section~\ref{results1}): instead of diverse metaphors, it repeats nearly the same metaphor across generations, indicating a narrow use of cultural knowledge.
Second, cultural context significantly reshapes conceptual relationships in the model's embedding space (Section~\ref{results2}): some metaphors cluster more tightly or spread farther apart compared to others, revealing that the model's internal notion of how concepts relate depends on the prompted culture. 
Third, the culture-neutral \textit{Default} setting is frequently \textit{not neutral} but instead aligns more closely with a Western (specifically U.S.) conceptual framing (Section~\ref{results3}).
Taken together, these results suggest that merely changing the prompt to invoke a culture does not guarantee truly culturally distinct reasoning.

\subsection{Implications for Co-Creative Use of LLMs}
Our findings have three key implications for using LLMs for co-creativity in cultural-specific settings.

While LLMs are often positioned as global creative partners, our results suggest that cultural prompting may provide only a surface-level adaptation. When a model defaults to a Western conceptual framework even under non-Western cultural prompts, it risks producing outputs that feel inauthentic or reductive, potentially homogenizing creative expression rather than enriching it. This limitation is particularly relevant for multicultural applications such as education, creative writing, and collaborative ideation. In such settings, a model that cannot meaningfully represent diverse cultural perspectives may function more as a translation or paraphrasing tool than as a genuine creative collaborator. Rather than introducing new viewpoints, it may repeatedly steer users toward a narrow set of familiar metaphors.

The presence of cultural defaultism also raises broader concerns for responsible AI design. It suggests that dominant cultural patterns in training data can shape the model’s implicit baseline, and that prompt engineering alone may be insufficient to address this issue. Our geometric analysis offers one approach to auditing such biases by examining how models organize abstract concepts, rather than focusing solely on surface-level content or harmful outputs. Supporting culturally inclusive co-creativity may therefore require both richer training data and evaluation methods that explicitly measure cultural variation in model representations.

\subsection{Limitations and Future Work}
This study has several limitations that point to directions for future work. 
First, we focus on metaphor generation as a probe for cultural reasoning. While metaphors capture important aspects of conceptual structure, they represent only one dimension of cultural worldview.
Second, all outputs were generated in English to enable direct semantic comparison. Although this avoids translation confounds, it may obscure language-specific expressions and culturally meaningful nuances that do not translate well.
Third, our preliminary analysis is limited to a single model, and future work could systematically evaluate a broader range of LLMs.
Finally, our algorithmic analyses abstract away pragmatic and stylistic differences between metaphors. Combining these analyses with human evaluation could provide a more complete understanding of how LLMs reason across cultures.

\bibliography{aaai2026}

\end{document}